\documentclass[letterpaper, 10 pt, conference]{ieeeconf}  

\IEEEoverridecommandlockouts                              

\usepackage{graphicx} 
\usepackage{epsfig} 
\usepackage{times} 
\usepackage{amsmath} 
\usepackage{amssymb}  
\usepackage{biblatex}
\usepackage{booktabs}
\usepackage{placeins}
\usepackage{multicol}
\usepackage{multirow}
\usepackage{graphicx}
\usepackage{textcomp}
\usepackage[dvipsnames]{xcolor}
\usepackage{textcase}
\usepackage[tablename=TABLE]{caption}
\let\labelindent\relax
\usepackage{enumitem}
\usepackage{diagbox}
\usepackage{caption}
\usepackage{float}
\usepackage{hyperref}

\title{\LARGE \bf
\textcolor{black}{PREGAN: Pose Randomization and Estimation for Weakly Paired Image Style Translation}
}

\author{Zexi Chen, Jiaxin Guo, Xuecheng Xu, Yunkai Wang, Yue Wang, Rong Xiong
\thanks{Zexi Chen, Jiaxin Guo, Xuecheng Xu, Yunkai Wang, Yue Wang, Rong Xiong are with the State Key Laboratory of Industrial Control Technology and Institute of Cyber-Systems and Control, Zhejiang University, Zhejiang, China. Yue Wang is the corresponding author {\tt\small wangyue@iipc.zju.edu.cn}.}}

\begin{document}

\maketitle
\thispagestyle{empty}
\pagestyle{empty}

\begin{abstract}
Utilizing the trained model under different conditions without data annotation is attractive for robot applications. \textcolor{black}{Towards this goal, one class of methods is to translate the image style from another environment to the one on which models are trained. In this paper, we propose a weakly-paired setting for the style translation, where the content in the two images is aligned with errors in poses. These images could be acquired by different sensors in different conditions that share an overlapping region, e.g. with LiDAR or stereo cameras, from sunny days or foggy nights. We consider this setting to be more practical with: (i) easier labeling than the paired data; (ii) better interpretability and detail retrieval than the unpaired data. To translate across such images, we propose PREGAN to train a style translator by intentionally transforming the two images with a random pose, and to estimate the given random pose by differentiable non-trainable pose estimator given that the more aligned in style, the better the estimated result is. Such adversarial training enforces the network to learn the style translation, avoiding being entangled with other variations.} Finally, PREGAN is validated on both simulated and real-world collected data to show the effectiveness. Results on down-stream tasks, classification, road segmentation, object detection, and feature matching show its potential for real applications.  {\small \href{https://github.com/wrld/PRoGAN}{\texttt{Code is available here}}.}
\end{abstract}

\section{Introduction}
\label{sec:INTRODUCTION}

\textcolor{black}{Domain transferring endows the robots with higher-level of understanding of the environment and helps robots to adapt to different scenarios and conditions by eliminating these differences within a common domain. Therefore, tasks trained on or configured on the common domain can be easily carried forward on other scenarios with such transfer learning, e.g. image style translation. This setting is extremely helpful for multiple robotics tasks, such as transferring a detector \cite{sun2006road} trained in sunny days to rainy nights as well as place recognition \cite{tang2020adversarial}, and semantic segmentation \cite{garcia2018survey} across domains.}

    \begin{figure}[t]
        \centering
            \includegraphics[width=\linewidth]{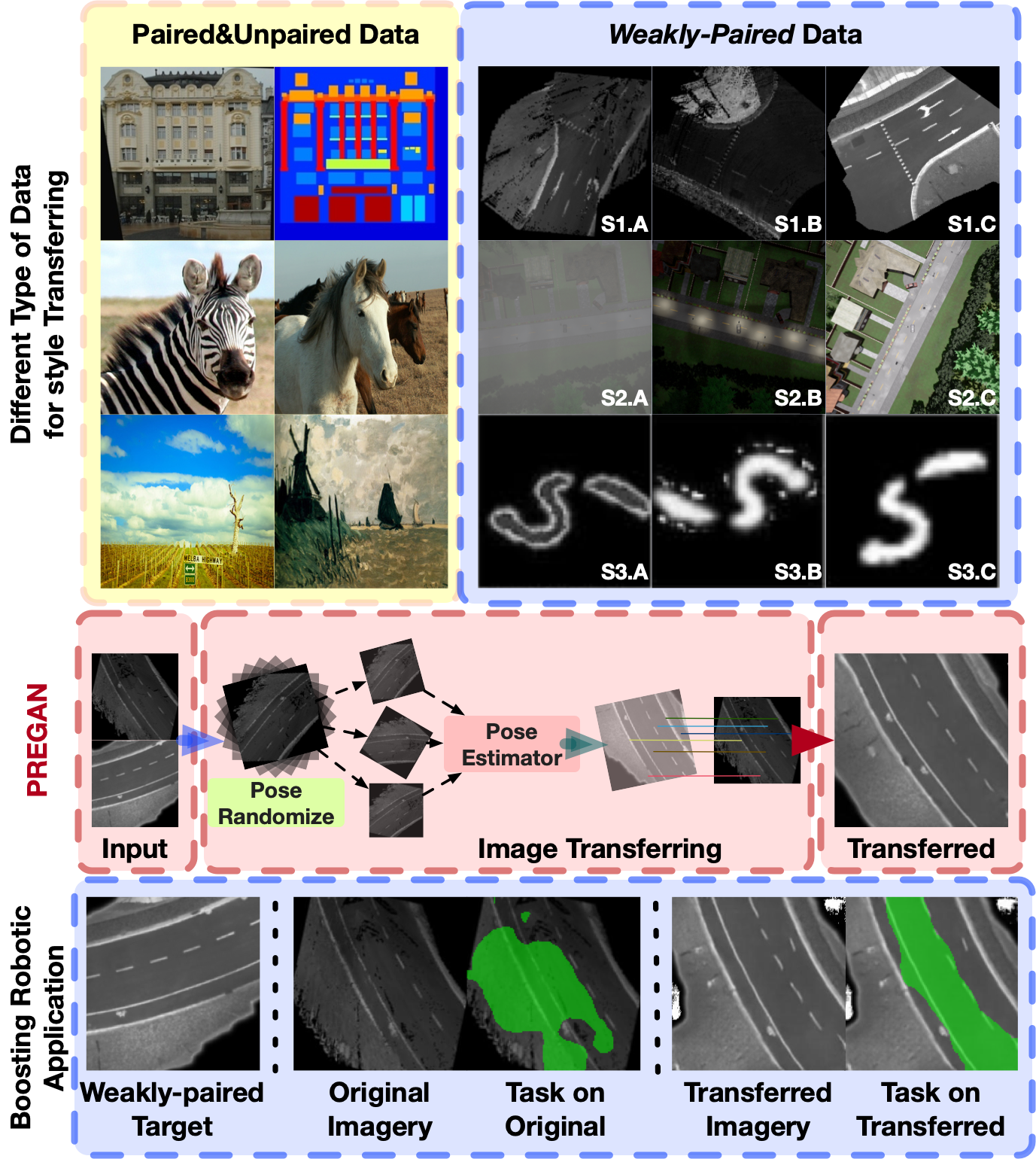}
        \caption{\textbf{Top Left} is a combination of widely adopted paired \& unpaired dataset for GANs' benchmarks. They are \textit{Facades}, \textit{Horse2Zebra}\, and \textit{Monet2Photo} from top to bottom with a consensus that objects are upright in the gravity direction. \textbf{Top Right} is the weakly-paired dataset that we conduct our experiments on, and they are obtained from/by: \textcolor{black}{\textit{WeakPair} dataset(S1\&S2), and transforming \textit{MNIST}(S3)}. \textbf{Middle} shows how PREGAN deals with weakly-paired data. \textbf{Bottom Left:} Weakly paired images for the style transfer training. \textbf{Bottom Middle:} Original imagery and task on it. \textbf{Bottom Right:} Transferred imagery by PREGAN and correspondingly succeeded task.}
        \label{fig:teaser}
        \vspace{-15pt}
    \end{figure}

 \textcolor{black}{Currently, in the image style translation, most works focus on either fully paired image translation or totally unpaired image translation. In the paired image translation, the contents for the two images from two domains are exactly the same. To deal with this task, \textsc{Pix2Pix}\cite{pix2pix2017} proposes a generative adversarial network (GAN) based solution and achieves good performance. However, in most robotic applications, it is not practical to obtain exactly paired data. To solve this, \cite{CycleGAN2017,choi2018stargan,huang2018munit} emerged with the ability of training style transferring models with unpaired images. The main idea of these works is to disentangle the content and style in the feature space, so that content and style features can be assembled. The disadvantages are obvious: (i) the boundary between the learned content and style can be vague, leading to the failure of translation as these unsupervised methods discard accuracy that comes with supervision; (ii) massive usage of the cyclic-consistency leads to the difficult training. }
 
 \textcolor{black}{Nevertheless, there is one exploitable area between paired and unpaired data that can close the gap above. \textit{We argue that in the robotic tasks, with all types of sensors, even though it is hard to collect exactly paired data, it is feasible to collect data with large overlap, which we name as \textbf{\textit{weakly-paired}} data.} Given the definition of \textit{weakly-paired} data, a style translator trained on it is a perfect match for boosting robotics tasks among different domains.}

 \textcolor{black}{To solve weakly-paired image style translation, in this paper, we propose Pose Randomization and Estimation GAN (PREGAN).} According to \cite{goodfellow2014generative}, the equilibrium of GAN is a match between the distribution of the generated data and that of the target data.  \textcolor{black}{By utilizing pose injection and pose recovery as domain randomization together with self-supervision, the generative adversarial process is expected to be insensitive to the pose and focuses on the style learning. Two incremental phases are proposed: \textbf{BAISC}: ``Pose Randomization'' containing random pose injection to the original image alone and pose recovery by estimator. \textbf{FULL}: introducing ``Self-supervision'' together with the BASIC version so that a further constraint on the content invariance is achieved.} In the experiments, PREGAN outperforms the unpaired data based methods on classification, segmentation, detection and feature points matching(Fig.\ref{fig:teaser}). The contributions can be summarized as
\begin{itemize}[leftmargin=*]
    \item  A new task is proposed as \emph{weakly-paired image style translation}, which tolerates reasonable pose error injected in the conventional paired data task, thus making it much more practical for robotic applications.
    \item  \textcolor{black}{PREGAN is proposed to solve the weakly-paired image style translation problem by artificially occupying the pose distribution of the two weakly-paired inputs with random pose and utilize the estimation consistency of the generated relative pose as hard self-supervised loss so that the content invariance can be better achieved.}
    \item A new weakly-paired dataset \emph{WeakPair} \cite{WeakPair} built in Carla \cite{dosovitskiy2017carla} and real-world is released for researches and this dataset supports the experiments of the proposed method, which demonstrates superior performance in 4 typical robotic tasks.
\end{itemize}

\section{Related work}
\label{sec:Related Work}

\subsection{Image to Image Translation}
\label{subsec:Image to Image Translation}
In computer vision, style translation problems are general tasks that aim to translate an input image from the original domain to the target domain. GANs have achieved excellent results in translation, without any hand-crafted labels.

Many models have been proposed for translating paired images. Introduced by Zhu et al.\cite{pix2pix2017}, \textsc{Pix2Pix} is a classical framework for paired image-to-image translation, which utilizes condition GAN to learn a mapping representation from input to output images. Some other methods also tackle the unpaired image translation problems with the goal of translating images from the original domain to the target domain without any alignments. \textsc{CycleGAN}\cite{CycleGAN2017} utilizes a cycle framework, of which the cycle-consistency loss provides a regularization to prevent generators from excessive hallucinations and mode collapse. \textsc{UNIT}\cite{liu2017unsupervised} assumes a shared-latent space and proposes an unsupervised image-to-image translation framework based on Coupled GANs. \textsc{StarGAN}\cite{choi2018stargan} is proposed to perform translations for multiple domains using only a single model, utilizing a mask vector method to control all available domain labels. \textsc{UGATIT}\cite{kim2019u} incorporates a new attention module and AdaLIN for unsupervised translation. \textsc{DRIT++}\cite{DRIT_plus} is presented to generate diverse outputs without paired training images based on disentangled representation. \textsc{MUNIT}\cite{huang2018munit} assumes that the image representation can be decomposed into a content code that is domain-invariant, and a style code that captures domain-specific properties. \textcolor{black}{\textsc{CrDoCo}\cite{CrDoCo} introduces a cross-domain consistency loss to capture pixel-level domain shifts that are critical to dense prediction. \textsc{CyCaDa}\cite{Hoffman_cycada2017} adapts representation at both pixel-level and feature-level which enhances cyclic consistency.}

 \textcolor{black}{These methods are state-of-the-art either in fully-paired data or unpaired data, however, none of them are capable of translating images who are highly entangled in pose and style, say, weakly-paired. Comparing to these methods, the proposed PREGAN solves the entanglement problem by introducing domain randomization and estimation together with cycle consistency.}

\subsection{Self-supervised Learning}
\label{subsec:Self-supervised Learning}
Self-supervised (SS) learning is a powerful approach for representation learning using unlabeled data. SS generally involves learning from tasks designed to resemble supervised learning in the way that ''labels" can be created from the data itself without human labor. \cite{gidaris2018unsupervised} utilizes input image rotations by four certain degrees and train the model with the 4-class classifier to recognize the four rotations. The pretext task offers a powerful supervisory signal for semantic feature learning.
    \begin{figure*}[t]
        \centering
            \includegraphics[width=0.9\textwidth]{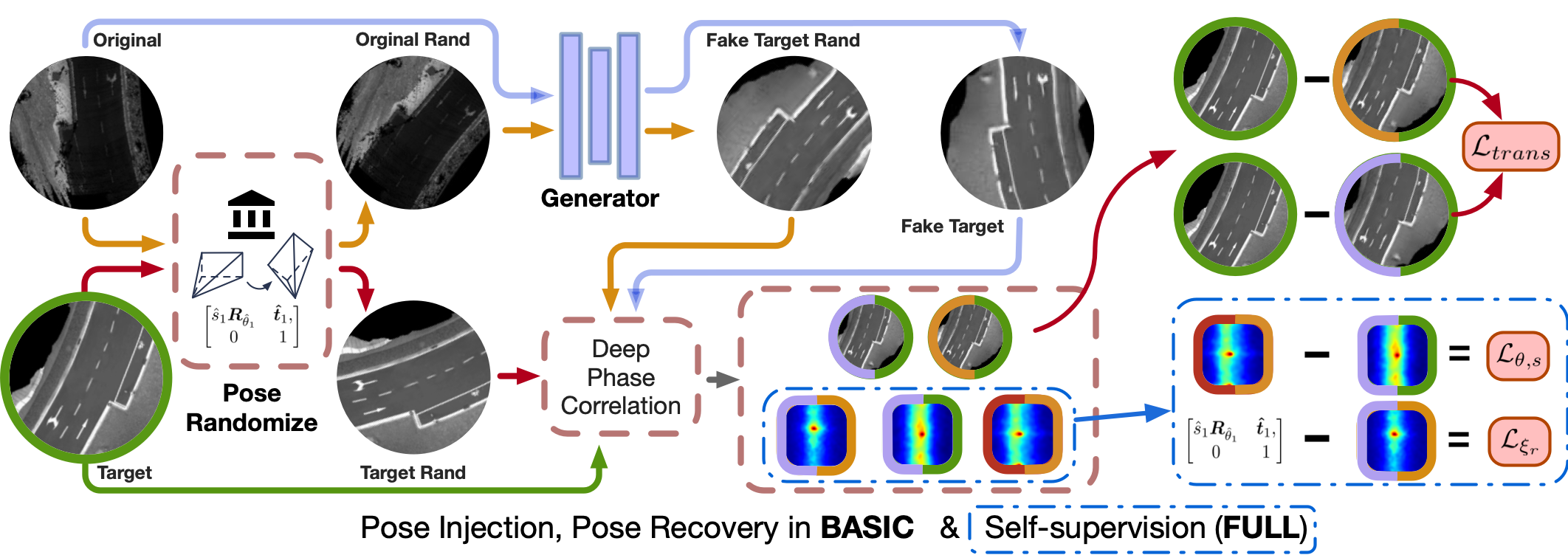}
        \caption{ \textcolor{black}{\textbf{BASIC} version is shown as contents outside the blue box, inside of which is the self-supervision for the \textbf{FULL} version. \textbf{BASIC:} ``Pose Randomization'': pose injection to $Original$ alone and pose recovery for $FakeTarget$ and $FakeTargetRand$. The image surrounded by purple and green(or orange and green) is the pose transformed $FakeTarget$(or $FakeTargetRand$) with respect to the estimated pose with $Target$. Auxiliary losses in this version is the pose estimation and recovery loss: $\mathcal{L}_{trans}$. \textbf{FULL:} Introduce self-supervision to the basic version. Heatmaps are the phase correlations of the two images from the two surrounding colors. Auxiliary Losses for self-supervision are $\mathcal{L}_{\theta,s},\mathcal{L}_{\xi_r}$.}}
        \label{fig:netstructure}
        \vspace{-15pt}
    \end{figure*}
However, during a GAN training process, it is often unstable and easily forgets previous tasks when searching for a Nash equilibrium in a high-dimensional space. A number of approaches based on SS have been proven successful for GANs training. Tran et al. \cite{tran2019improved} aim to improve GANs by applying SS learning via the geometric transformation on input images and assign the pseudo-labels to these transformed images. Chen et al. propose the self-supervised GAN \cite{chen2019self}, adding auxiliary rotation loss as a self-supervised loss to the discriminator.

 \textcolor{black}{Compared to existing self-supervisions introduced to GAN that mostly counter with the problem of the forgetfulness of the discriminator by a hand crafted classifier, our method utilizes a continuous pose estimator to help the pose randomization and the generator so that PREGAN can easily disentangle pose and style.}

\subsection{Domain Randomization}
\label{subsec:Domain Randomization}
Domain randomization (DR) is a complementary class of methods for domain adaptation, aiming to randomize parts of the domain to which the model should not be sensitive. Tobin et al. \cite{tobin2017domain} utilize DR for addressing the reality gap. They randomize the simulator to expose the model to different environments at training time. \cite{yue2019domain} is proposed to randomize the synthetic images with styles of real images in terms of visual appearance using auxiliary sim-to-real datasets.

 \textcolor{black}{Inspired by the domain randomization, this paper introduces pose randomization to the training process with the purpose of weakening the effect of the misaligned pose. Furthermore, we introduces pose estimator for pose recovery of the original image so that the effect of the pose randomization is further enhanced.}

\section{PREGAN for weakly-paired image style translation}
\label{sec:Methodology}

A typical example for paired, unpaired and the proposed weakly-paired data is shown in Fig. \ref{fig:teaser}. One can see that for paired data, the two images with different style, have perfect aligned content. For unpaired one, the two images are unrelated in data. For weakly-paired data, contents are similar but misaligned in pose.  \textcolor{black}{Since robots carry multi-sensors to perceive geometric measurements, coarsely alignment of the images with different styles is reasonable and therefore, we consider that such a setting is not trivial and worth to be dealt with.}

We formally state the weakly-paired image style translation as a generative process of target image $I_{t}$ using the original image $I_{o}$:
\begin{equation}\label{wpm}
I_{t} = T(f(I_{o}),\xi)
\end{equation}
where $f$ and $T$ are the style translation process, and the image geometric transformation process, $\xi$ is the relative pose between the two images, which parameterizes $T$. Note that relative pose in weakly-paired data is in $\mathbb{SIM}(2)$
\begin{gather}
\xi = \begin{pmatrix}
s\boldsymbol{R}_{\theta} & \boldsymbol{t} \\
0 & 1
\end{pmatrix} \in \mathbb{SIM}(2)
\label{eq: solution space}
\end{gather}
where $s \in \mathbb{R}^+$ is the scale, $\boldsymbol{R}_{\theta} \in \mathbb{SO}(2)$ is the rotation matrix generated by the heading angle $\theta$, and $\boldsymbol{t} \in \mathbb{R}^2$ is the displacement vector. \textcolor{black}{\textbf{The relative pose $\xi$ between $I_{o}$ and $I_{t}$ in this problem remains unknown.}}

The aim of \textit{weakly-paired} data style translation is to learn a network $\mathcal{N}_{t}$ to approximate $f$ from dataset $\{I_{o},I_{t}\}$, so that \textcolor{black}{the} original image can be translated using $\mathcal{N}_{t}$. The challenge of this problem is to keep the learned function $\mathcal{N}_{t}$ only translating style. Simply learning a network $\mathcal{N}_{t}$ between the two images obviously leads to a process entangled with style, pose and content. On the other hand, regarding the image pair as unpaired does not utilize the information of the bounded $\xi$, leaving a large improvement margin. \textcolor{black}{Putting aside the fully supervised method for paired images, the unsupervised methods are easily confused by the complicated pose of weakly-paired images and misses the benefits of pixel-wise correspondence. We argue that a new method considering the high entanglement of pose and style is urgently needed.}

\subsection{Pose Randomization}

\textcolor{black}{Inspired by the domain randomization which aims to leverage the insensitivity to the randomized domain, we propose pose randomization to keep the network focusing on style transferring while training on the weakly-paired data. Given the two images $\{I_{o},I_{t}\}$  of the same scene that are different in style and pose, the main idea is to eliminate the entanglement by intentionally injecting relative pose $\xi_r$ to the original domain images $I_o$ and recover these images to the pose of the target with an estimator. Theoretically, the original image $I_o$ and the randomized original image $I_{o,\xi_r}$ make no difference when comparing to the target image $I_{t}$ at the pose distribution level since they are misaligned but related. By taking this into account, the distribution of image differences between the three domain are only induced by style changes and we can train a $\mathcal{N}_{t}$ to capture this process.}

\noindent \textsc{\textbf{Transformed image distribution match:}} \textcolor{black}{$T$ and $f$ are commutative, then we have}
\begin{equation}\label{comm}
I_{t} = f(T(I_{o},\xi)) =f(I_{o,\xi}) 
\end{equation}
where $I_{o,\xi}$ is $I_o$ transformed with $\xi$. Note that we do not really have such images as $\xi$ for each pair is unknown. However, as we roughly know the distribution of $\xi$ according to the robot sensors, we can randomize $\xi$ with $\xi_r$ to generate a set of $\{I_{o,\xi_r}\}$, \textcolor{black}{leading to a distribution match of generated images $I_{o,\xi_r}$ and the real transformed images $I_{o,\xi}$}
\begin{equation}\label{density}
p(I_{o,\xi}) \approx p(I_{o,\xi_r})
\end{equation}
where $p(\cdot)$ is the pose's probability density of $\cdot$. \textcolor{black}{By (\ref{density}), we agree that the empirical distribution of poses in a set of $\{I_{o,\xi_r}\}$ is subjected to the true distribution of $I_{o,\xi}$ when $\xi_r$ is in a reasonable range.}

\noindent \textsc{\textbf{Target image distribution match:}} We then apply a neural network $\mathcal{N}_{t}$ to transfer the style of $I_{o,\xi_r}$ to $\tilde{I}_{t}$
\begin{equation}\label{transfer}
\tilde{I}_{t} = \mathcal{N}_{t}(I_{o,\xi_r}),~~I_{o,\xi_r} \sim p(I_{o,\xi_r}),
\end{equation}
\textcolor{black}{where $I_{o,\xi_r}$ is subjected to the distribution of $p(I_{o,\xi_r})$.} Recalling (\ref{comm}), we can approximate the generative process of $I_t$ with (\ref{transfer}) in the sense of probability distribution. 

Since GAN \cite{goodfellow2014generative} provides a tool for matching the distribution of generated images and the target images, \textcolor{black}{we utilize it to train the match of $p(\tilde{I}_{t})$ and $p(I_t)$. Specifically, the discriminator loss $\mathcal{L}_{D}$ with PatchGAN \cite{pix2pix2017} is designed as predicting $\Tilde{I_{t}}$ to be False and $I_{t}$ to be true:}
    \begin{equation}
            \mathcal{L}_{D} =  P_{\Tilde{I_{t}}}^{F} + P_{I_{t}}^{T},
    \label{eq: DLOSS}
    \end{equation}
and the generator loss $\mathcal{L}_{G}$ is designed as \textcolor{black}{predicting $\Tilde{I_{t}}$ to be true:}
    \begin{equation}
            \mathcal{L}_{G} = P_{\Tilde{I_{t}}}^{T},
    \label{eq: DLOSS}
    \end{equation}
where $P_{\cdot}^{T}$ and $P_{\cdot}^{F}$ stands for predicting $\cdot$ to be true and false respectively.

\noindent \textsc{\textbf{Relative pose generator:}} Based on the GAN introduced above, $\mathcal{N}_{t}$ is learned to approximate $f$ to transfer the style of the image. We further design a solver to estimate the exact relative pose $\hat{\xi}$ between a pair of data. Therefore, we can follow (\ref{wpm}) to first change the style as $\Tilde{I}_t = f(I_o)$, then use the estimated $\hat{\xi}$ to generate the final $\check{I}_t$.

To estimate the relative pose $\hat{\xi}$, we leverage the deep phase correlation (\textsc{DPC})\cite{chen2020deep}, which results in a distribution of relative rotation and scale changes $p(\hat{\xi}^{\hat{\theta}_{1},\hat{s}_{1}})$ by 
\begin{gather}
p(\hat{\xi}^{\hat{\theta}_{1},\hat{s}_{1}}) = \mathfrak{C}(\mathfrak{L}(\mathfrak{F}(\Tilde{I}_{t})),\mathfrak{L}(\mathfrak{F}(I_{t})))
\label{eq: CorrMAP}
\end{gather}
where $\mathfrak{F}$ is the discrete Fourier Transform, $\mathfrak{L}$ is the log-polar transform and $\mathfrak{C}$ is the phase correlation solver. \textcolor{black}{Fourier Transformation $\mathfrak{F}$ transforms images into the Fourier frequency domain of which the magnitude has the property of translational insensitivity, therefore the rotation and scale are decoupled with displacements and are represented in the magnitude. Log-polar transformation $\mathfrak{L}$ transforms Cartesian coordinates into log-polar coordinates so that such rotation and scale in the magnitude of Fourier domain are remapped into displacement in the new coordinates, making it solvable with the afterward phase correlation solver. Phase correlation solver $\mathfrak{C}$ outputs a heatmap indicating the displacements of the two log-polar images, which eventually stands for the rotation and scale of the two input images $\Tilde{I}_{t}$ and $I_{t}$. To make the solver differentiable, we use expectation as the estimation of $\hat{\xi}$.} Thus the rotation $\hat{\theta}_{1}$ and scale $\hat{s}_{1}$ are 
\begin{gather}
(\hat{\theta}_{1},\hat{s}_{1}) = \mathop{\mathbb{E}}(p(\hat{\xi}^{\hat{\theta}_{1},\hat{s}_{1}})),\\
 \bar{I_{t}} =  \begin{bmatrix}
\hat{s}_{1}\boldsymbol{R}_{\hat{\theta}_{1}} & 0, \\
0 & 1
\end{bmatrix} \Tilde{I_{t}}
\label{eq: Rotate}
\end{gather}
where $\mathop{\mathbb{E}}(\cdot)$ is the expectation of $\cdot$.  We rotate and resize $\Tilde{I}_{t}$ referring to $\hat{\theta}_{1}$ and $\hat{s}_{1}$ with the result of $\bar{I}_{t}$ and calculate the relative displacement distribution $p(\hat{\xi}^{\hat{\textbf{t}}_{1}})$ between  $\bar{I}_{t}$ and $I_{t}$:
\vspace{-5pt}
    \begin{gather}
   p(\hat{\xi}^{\hat{\textbf{t}}_{1}}) = \mathfrak{C}(\bar{I_{t}},I_{t}),\\
    \hat{\boldsymbol{t}_{1}} = \mathop{\mathbb{E}}(p(\hat{\xi}^{\hat{\textbf{t}}_{1}})).
    \label{eq: Transform}
    \end{gather}
Finally, we arrive at $\check{I}_t$ by\textcolor{black}{
    \begin{gather}
    \check{I_{t}} =  \begin{bmatrix}
    1 & \hat{\boldsymbol{t}}_{1}, \\
    0 & 1
    \end{bmatrix}  \bar{I_{t}} 
    = 
    \begin{bmatrix}
    \hat{s}_{1}\boldsymbol{R}_{\hat{\theta}_{1}} & \boldsymbol{\hat{t}}_{1}, \\
    0 & 1
    \end{bmatrix} \Tilde{I}_{t}.
    \label{eq: Translated}
    \end{gather}}
By now, the generated image has the process
\begin{equation}\label{gen}
\check{I}_t = T(\mathcal{N}_{t}(I_o),\xi).
\end{equation}
Obviously, $\check{I}_t$ should be exactly the same with $I_{t}$. Moreover, we can also apply the same procedure to $I_{o,\xi_r}$ to generate $\check{I}_{t,\xi_r}$, which should also be the same to \textcolor{black}{$I_{t}$}. Therefore, following the one-to-one generated target image, we can design a pixel level loss terms as
\textcolor{black}{\begin{equation}\label{l1}
\mathcal{L}_{trans} = \lVert I_{t}-\check{I}_t\lVert_{1}  + \lVert I_{t}-\check{I}_{t,\xi_r}\lVert_{1},
\end{equation}}
where $\lVert \cdot \lVert_{1}$ is the L1 norm of $\cdot$.

\noindent \textsc{\textbf{Joint loss and training:}} Before introducing the loss function, one cyclic manner is adopted in the PREGAN to avoid model collapse. An $\mathcal{N}_{o}$ is leveraged to translate $I_{t}$ back to a fake $I_{o}$. Now $\hat{I_{t}}$ will goes through $\mathcal{N}_{o}$ after it got transformed from $\mathcal{N}_{t}$ and becomes a $I_{o,c}$. 
    \begin{equation}
        I_{o,c} = \mathcal{N}_{o}[\mathcal{N}_{t}(I_{o})]
        \label{Iocycle}
    \end{equation}
At this point, the \textbf{BASIC} version of PREGAN is explicitly introduced and the losses in the training include one from the generator and another one from the discriminator. Losses $\mathcal{L}_{G}$ calculated in the generator includes losses from CGAN in predicting the desired identity of $\Tilde{I}_{t}$ and $\check{I}_{t}$ with respect to $I_{t}$, the enhanced $L1$ loss between the set $\check{I}_{t},I_{t}$ and $\check{I}_{t,\xi_r},I_{t,\xi_r}$ and a cycle loss between $I_{o,c}$ and $I_{o}$:
    \begin{equation}
    \begin{split}
    & \mathcal{L}_{cycle} = \lVert I_{o}-I_{o,c}\lVert_{1},
    \label{eq: PartLOSS}
    \end{split}
    \end{equation}
    \begin{equation}
            \mathcal{L}_{G} = \mathcal{L}_{trans} + \mathcal{L}_{cycle} + P_{\Tilde{I}_{t}}^{T} + P_{\check{I}_{t}}^{T}.
    \label{eq: GLOSS}
    \end{equation}
The discriminator loss is constructed by predicting the true identity of $\Tilde{I}_{t}$, $\check{I}_{t}$, $\check{I}_{t,\xi_r}$ and $I_{t}$:
    \begin{equation}
            \mathcal{L}_{D} = P_{\Tilde{I}_{t}}^{F} + P_{\check{I}_{t}}^{F} + P_{\check{I}_{t,\xi_r}}^{F} + P_{ I_{t}}^{T},
    \label{eq: DLOSS}
    \end{equation}
and the total loss $\mathcal{L}_{All}$ is the sum of $\mathcal{L}_{G}$ and $\mathcal{L}_{D}$:
    \begin{equation}
    \mathcal{L}_{PREGAN_{BASIC}} = \mathcal{L}_{G} + \mathcal{L}_{D}
    \label{eq: BASICALLLOSS}
    \end{equation}

\subsection{Self-supervision Tasks}

During the training of PREGAN, there are intermediate results whose relative poses are known, which can thus be utilized as a self-supervision to further constrain the network, improving the final performance. Elaborations can be found in the Appendix.

\noindent \textsc{\textbf{Randomized pose as self-supervision:}} In pose randomization, we intentionally inject the noisy pose $\xi_r$ to transform the images. This pose is exactly known, and can be utilized as a self-supervision. As mentioned in (\ref{gen}), we have $\check{I}_t$ and $\check{I}_{t,\xi_r}$ generated by $N_{t}$ and $T$. Their relative angle is $\xi_r$. Motivated by this equality, we apply the DPC in (\ref{eq: Rotate}) and (\ref{eq: Translated}) to $\check{I}_t$ and $\check{I}_{t,\xi_r}$, resulting in the distribution of $p(\hat{\xi}_r)$, \textcolor{black}{which is supervised by the KLD between the Gaussian Blurred one-peak distribution centering at $\xi_r$}
\vspace{-5pt}
\begin{equation}\label{eq: kldxir}
\mathcal{L}_{\xi_r} = KLD(p(\hat{\xi}_r),\textbf{1}_{\xi_r})
\end{equation}
where $\textbf{1}_{\cdot}$ indicates for the one-peak distribution centering at $\cdot$. For implementation, we only use the first stage of DPC to estimate rotation, upon which the loss term is built for simplicity. 

\noindent \textsc{\textbf{Relative pose equality as self-supervision:}} Another supervision is built between the relative pose estimation from $(I_t,\Tilde{I}_t)$, and $(I_{t,\xi_r},\Tilde{I}_{t,\xi_r})$. Note that only the rotation part of the two estimation result should be equal. The translation part is affected by the noisy pose $\xi_r$. Therefore, we use the intermediate DPC output of $p(\hat{\theta}_{1},\hat{s}_{1})$ as in (\ref{eq: Rotate}), and $p(\hat{\theta}_{\xi_r},\hat{s}_{\xi_r})$, to measure their similarity, leading to a KLD based loss term as
\vspace{-5pt}
\begin{equation}\label{eq: kldsr}
\mathcal{L}_{\theta,s} = KLD[p(\hat{\theta}_{1},\hat{s}_{1}),p(\hat{\theta}_{\xi_r},\hat{s}_{\xi_r})]
\end{equation}

With all the loss terms introduced above (\ref{eq: BASICALLLOSS}), (\ref{eq: kldxir}) and (\ref{eq: kldsr}), we finally present the joint loss as
\vspace{-5pt}
\begin{equation}\label{Total}
\mathcal{L}_{PREGAN_{FULL}} =\mathcal{L}_{PREGAN_{BASIC}}+\mathcal{L}_{\theta,s}+\mathcal{L}_{\xi_r}
\end{equation}
Upon which, we name the resultant network as \textbf{FULL} version of PREGAN.

\subsection{Implementation Details}
To achieve high-resolution image translation, we adopt the network from \cite{johnson2016perceptual}, which contains Resnet blocks for generator network. To realize the classification for each patch, we adopt $70 \times 70$ PatchGAN \cite{pix2pix2017} for the discriminator network, which has fewer parameters and can work on arbitrary size images. \textcolor{black}{The random pose injections in ``Pose Randomization'' and ``Self-supervision'' are independent: (i) pose injection is applied to the original image alone in ``Pose Randomization'' and (ii) is applied to both of the input images in ``Self-supervision''. The difference between it and data augmentation is further discussed in the Appendix.} All networks are trained from scratch with batch normalization and the learning rate of $1.5\times10^{-5}$.

\section{{Experiments: Dataset And Setup}}
\label{sec:Experiments and Results}

\textcolor{black}{Our approach is evaluated on both simulation and real-world datasets including four different tasks: classification, object detection, feature matching and road segmentation. Classification is conducted on the \textit{MNIST} dataset, while the rest of the tasks are conducted on the proposed \textit{WeakPair} dataset which contains simulation settings from CARLA, and real world settings collected by drone and the ground robot.}
        
\subsection{Dataset}
\label{subsec:Dataset}
\noindent \textbf{\textsc{\textit{WeakPair:}}} \textcolor{black}{The \textit{WeakPair} dataset is collected both in the autonomous driving simulator in four different weather conditions and in the real world with multiple sensors and perspectives.}
\begin{itemize}[leftmargin=*]
    \item \textbf{\textcolor{black}{Simulation in Carla (\textit{WeakPair(S)})}:} It is recorded in the overhead orthomosaic perspective and is leveraged in the paper whose demonstration in different weathers is shown in Fig. \ref{fig:teaser} where S2.A is the ``foggy" , S2.B is the ``night", S2.C is the ``sunny". Each task in this paper requiring this dataset has its competing methods trained with weakly-paired images to transfer adverse weathers to a sunny one in Town 1 and tests their corresponding performances in Town 2.
    \item \textcolor{black}{\textbf{Real world data (\textit{WeakPair(R)}):} It contains several different images pairs shown as follows and in this paper, our experiments on \textit{WeakPair(R)} are conducted on L2D and S2D:}
        \begin{enumerate}
        \item \textbf{L2D:}``LiDAR Local Map" to ``Drone's View";
        \item \textbf{L2Sat:}``LiDAR Local Map" to ``Satellite Map";
        \item \textbf{S2D:}``Stereo Local Map" to ``Drone's View";
        \item \textbf{S2Sat:}``Stereo Local Map" to ``Satellite Map".
        \end{enumerate}
\textcolor{black}{Tasks requiring this dataset in the paper has their competing methods trained to transfer LiDAR representations(Fig.\ref{fig:teaser} S1.B) and stereo images(Fig.\ref{fig:teaser} S1.A) to the drone's style(Fig.\ref{fig:teaser} S1.C).}
\end{itemize}

\noindent \textbf{\textsc{\textit{MNIST:}}} The \textit{MNIST} dataset is a collection of hand-written numbers for the task of classification collected by \cite{lecun1998gradient}. We make it a weakly-paired one by changing styles via kernel disruption and Gaussian blur, and transforming poses, shown in Fig. \ref{fig:teaser}(S3). 

For all dataset above, we constrain translations of both $x$ and $y$, rotation changes and scale changes of the two weakly-paired images in the range of $[-50, 50]$ pixels, $[0, \pi)$ and $[0.8, 1.2]$ respectively with images shapes of $256 \times 256$.

\subsection{Tasks and Metrics}
\label{subsec:Tasks and Metrics}

 \noindent \textbf{\textsc{Classification:}} Each method is trained with weakly-paired images(e.g. S3.A and S3.B in Fig. \ref{fig:teaser}). We train a classifier\cite{he2016deep} on the original S3.C and test the classification on the transferred images. We adopt average precision $\mathcal{AP}_{class}$ and the harmonic mean of recall and $\mathcal{AP}_{class}$, noted as MaxF1 to evaluate the domain transfer performance.
 
  \noindent \textbf{\textsc{Road segmentation:}} For the two datasets, we train the segmentation network\cite{teichmann2018multinet} on \textit{WeakPair(S)}'s sunny day and on \textit{WeakPair(R)}'s drone's perspective and test the road segmentation performance on the style transferred images. We evaluate the performance with  Intersection-over-Union(IoU) $\mathcal{P}_{road}$(the same measurement as $\mathcal{P}_{class}$), and the harmony mean(MaxF1) of $\mathcal{P}_{road}$ and $\mathcal{R}_{road}$.
    
 \noindent \textbf{\textsc{Object detection:}} Baselines as well as PREGAN is trained to transfer adverse weathers to a sunny one in \textit{WeakPair(S)}. The detector\cite{redmon2018yolov3} is trained on the sunny day and evaluates the style translation performance of each method by mean average precision (mAP) and harmony mean of precision and recall (F1).

 \noindent \textbf{\textsc{Feature matching:}} The feature matching task is carried out with SIFT\cite{lowe2004distinctive} since it is a perfect solution to find the matching features of two images. It is adopted to evaluate the performance of each method for the more the generated data is alike with the target one, the more the SIFT points will be. We extract top 50 feature points from SIFT and calculate the true positive matching number $\mathcal{N}_{SIFT}$. \textcolor{black}{Details of it can be found in the Appendix.}

\subsection{Comparative Methods}
\label{subsec:Comparative Methods}

Baselines in the experiment include unpaired data based methods, \textsc{CycleGAN}\cite{CycleGAN2017}, \textsc{DRIT}\cite{DRIT_plus}, \textsc{MUNIT}\cite{huang2018munit}, \textsc{StarGAN v2}\cite{choi2020stargan}, \textsc{UGATIT}\cite{kim2019u}, \textcolor{black}{\textsc{CyCaDa}\cite{Hoffman_cycada2017}} and paired data based method \textsc{Pix2Pix}\cite{pix2pix2017}. In classification task, all baselines above are compared, and \textsc{Pix2Pix} is especially involved in this task with precisely paired data. By conducting experiments on both weakly-paired data and paired-data in the \textit{MNIST} dataset, we verify the upper bound of our method comparing to \textsc{Pix2Pix} in paired data and the better performance of our work comparing to rest of the baselines in weakly-paired data. For the remaining tasks, only unpaired data based methods above are compared. \textcolor{black}{For fairness, we apply two ways of random pose injection((i)the original image alone,(ii)both input images) that exists in the paper separately to all the baselines. In the main paper, providing baselines with type(i) pose injection as data augmentation is demonstrated whereas results on providing baselines with type(ii) pose injection as data augmentation is in the Appendix.}

\section{{Experiments: Results}}
\subsection{MNIST Classification}
\label{subsec:MNIST Classification}

\begin{table}[t]
\centering
\textcolor{black}{\caption{Evaluation results of classification on \textit{MNIST} dataset. \textsc{Pix2Pix2(P)} is claimed as \textsc{Pix2Pix} trained on paired data and \textsc{Pix2Pix2(W)} for \textsc{Pix2Pix} trained on weakly-paired data. AP reports the average precision and ``MaxF1" reports the harmony mean.}
\label{table: classification}
\resizebox{\linewidth}{!}{%
\begin{tabular}{llllll}
\toprule
\textbf{Baselines} & \textsc{Original} & \textsc{CycleGan} & \textsc{UGATIT}      & \textsc{StarGAN}     & \textsc{DRIT}   \\ \hline
AP                 & 64.1     & 86.7     & 78.1        & 41.7        & 88.2   \\
MaxF1              & 57.1     & 84.2     & 73.9        & 38.7        & 77.3   \\
\textbf{Baselines} & \textsc{MUNIT}    & \textsc{CyCaDa}   & \textsc{Pix2Pix(W)} & \textsc{Pix2Pix(P)} & \textsc{PREGAN} \\ \hline
AP                 & 86.9    &  85.7        & 65.43       & \textcolor{red}{93.2}        & \textcolor{blue}{92.6}   \\
MaxF1              & 82.1     &   83.1       & 69.6        & \textcolor{red}{94.0}        & \textcolor{blue}{93.9}  \\ \bottomrule
\end{tabular}%
}
\vspace{-15pt}}
\end{table}
    
In this experiment, we evaluate the capability \textcolor{black}{boundary} of our approach where we assume that PREGAN trained on weakly-paired \textit{MNIST} should not outperform \textsc{Pix2Pix} trained on perfectly paired \textit{MNIST}. While each competing method is trained on weakly-paired data, baseline \textsc{Pix2Pix} is trained on both weakly-paired data and paired data separately. For the qualitative result, please consult to the Appendix. The quantitative result in TABLE. \ref{table: classification} denotes that classification by \cite{he2016deep} informs a similar quality when it is performed on the raw \textit{MNIST} data, the output of PREGAN, and the output of \textsc{Pix2Pix}. It shows that PREGAN's capability of handling weakly-paired images is almost similar to \textsc{Pix2Pix} ability in paired images. However, \textsc{Pix2Pix} trained on paired data(\textsc{Pix2Pix(P)}) gives the best performance while it acts terribly when being trained on weakly-paired data(\textsc{Pix2Pix(W)}). Therefore, we believe that \textsc{Pix2Pix} is no longer a suitable baseline for the following tasks as only weakly-paired data is available.

\subsection{Road Segmentation}
\label{subsec:Road Segmentation}
    \begin{figure}[t]
        \centering
        \includegraphics[width=0.9\linewidth]{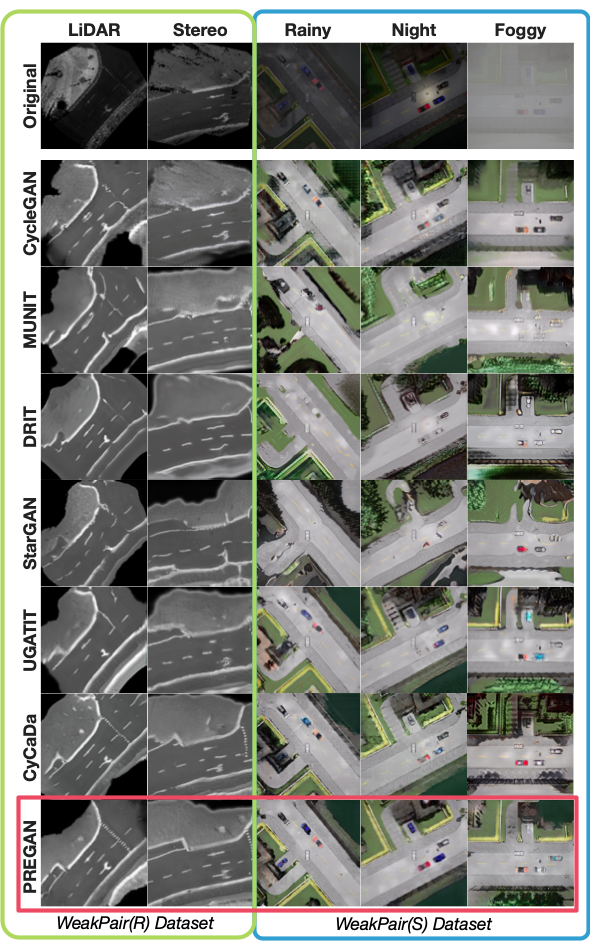}
        \vspace{-5pt}
        \caption{Qualitative comparison of competing methods in \textit{WeakPair(S)} and \textit{WeakPair(R)} dataset. The ``Original" is the input of each method and the ``Target" is the desired domain with weakly-paired geometrical relation. With good disentanglement, the output images of each method should only learns the style changes.}
        \label{fig:visualComparison}
        \vspace{-15pt}
    \end{figure}
In this experiment, we evaluate PREGAN on the task of road segmentation in two different datasets.

    \begin{table}[t]
    \textcolor{black}{
    \caption{Evaluation results of road segmentation on orthomosaic perspective in \textit{WeakPair(S)} dataset. The experiments is conducted on ``Foggy" $\rightarrow$ ``Sunny", ``Rainy" $\rightarrow$ ``Sunny", and ``Night" $\rightarrow$ ``Sunny". IoU(percentage) reports the segmentation performance of roads and the higher is better.}
    \label{table: WP Seg}
    \resizebox{\linewidth}{!}{%
    \renewcommand{\arraystretch}{1.1}
    \begin{tabular}{@{}lllllll@{}}
    \toprule
    \multicolumn{1}{c}{\multirow{2}{*}{\textbf{Baselines}}} &
      \multicolumn{2}{c}{Foggy(IoU)} &
      \multicolumn{2}{c}{Rainy(IoU)} &
      \multicolumn{2}{c}{Night(IoU)} \\ \cmidrule(l){2-7}
    \multicolumn{1}{c}{} &
      \multicolumn{1}{c}{AP} &
      MaxF1 &
      \multicolumn{1}{c}{AP} &
      MaxF1 &
      \multicolumn{1}{c}{AP} &
      MaxF1 \\ \cmidrule(r){1-1}
             Sunny(GT)  &   92.91 & 97.40 &  92.91 & 97.40 &  92.91  & 97.40 \\
             Original   &   65.73 & 77.54 &  23.26 & 37.74 &  30.17  & 37.65 \\
    \textsc{CycleGAN}   &   89.85 & 87.63 &  88.52 & 89.21 &  87.02  & 92.55\\
    \textsc{UGATIT}     &   87.52 & 84.66 &  81.93 & 82.07 &  88.19  & 91.44 \\
    \textsc{StarGANv2}  &   85.72 & \textcolor{blue}{89.57} &  88.24 & 89.29 &  \textcolor{blue}{88.50}  & \textcolor{blue}{92.96} \\
    \textsc{DRIT}       &   74.55 & 78.84 &  83.62 & 86.13 &  82.98  & 84.06 \\
    \textsc{MUNIT}      &   81.40 & 82.61 &  89.02 & 90.75 &  84.82  & 82.40 \\
    \textsc{CyCaDa}     &   \textcolor{blue}{90.52} & 88.66 &  \textcolor{blue}{91.02} & \textcolor{blue}{90.89} &  85.47  & 88.71 \\
    \textbf{Ours(PREGAN)} & \textcolor{red}{92.15} & \textcolor{red}{94.22} & \textcolor{red}{93.57} & \textcolor{red}{96.59} & \textcolor{red}{92.50} & \textcolor{red}{96.86} \\ \bottomrule
    \end{tabular}%
    }
    \vspace{-15pt}
    }
    \end{table}

    \begin{table}[t]
    \centering
    \textcolor{black}{\caption{Evaluation results on \textit{WeakPair(R)} Dataset of road segmentation. The experiments is conducted both on LiDAR map $\rightarrow$ Drones's Map and on stereo map $\rightarrow$ Drones's Map. IoU(percentage) reports the segmentation performance of roads and the higher is better.}
    \label{table: AG Seg}
    \vspace{0pt}
    \resizebox{0.7\linewidth}{!}{%
    \renewcommand{\arraystretch}{1.1}
    \begin{tabular}{@{}lllll@{}}
    \toprule
    \multicolumn{1}{c}{\multirow{2}{*}{\textbf{Baselines}}} &
      \multicolumn{2}{c}{S2A(IoU)} &
      \multicolumn{2}{c}{L2A(IoU)} \\ \cmidrule(l){2-5}
    \multicolumn{1}{c}{} &
      \multicolumn{1}{c}{AP} &
      MaxF1 &
      \multicolumn{1}{c}{AP} &
      MaxF1 \\ \cmidrule(r){1-1}
             Aerial(GT)            & 92.46  & 97.71 & 92.46 & 97.71 \\
             Original              & 54.67  & 53.59 & 16.12 & 28.26 \\
    \textsc{CycleGAN}              & 21.08  & 27.26 & 26.74 & 40.61 \\
    \textsc{UGATIT}                & 69.63  & 74.29 & 20.21 & 32.48 \\
    \textsc{StarGANv2}             & \textcolor{blue}{78.92}  & \textcolor{blue}{77.55} & 29.98 & 39.11 \\
    \textsc{DRIT}                  & 51.94  & 52.00 & 19.91 & 32.02 \\
    \textsc{MUNIT}                 & 39.27  & 40.08 & 26.10 & 37.82 \\
    \textsc{CyCaDa}                & 62.86  & 65.33 & \textcolor{blue}{41.09} & \textcolor{blue}{49.87} \\
    \textbf{Ours(PREGAN)} & \textcolor{red}{92.06} & \textcolor{red}{94.09} & \textcolor{red}{92.10} & \textcolor{red}{95.21} \\ \bottomrule
    \end{tabular}%
    }
    \vspace{-15pt}
    }
    \end{table}
\noindent\textbf{\textit{WeakPair(S)} \textsc{Dataset}:} Fig. \ref{fig:visualComparison} provides a qualitative comparison of the competing methods. We observe that our method synthesizes images with a higher visual quality and a better disentanglement compared to the baseline models.
 \begin{figure*}[t]
    \centering
    \includegraphics[width=0.7\textwidth]{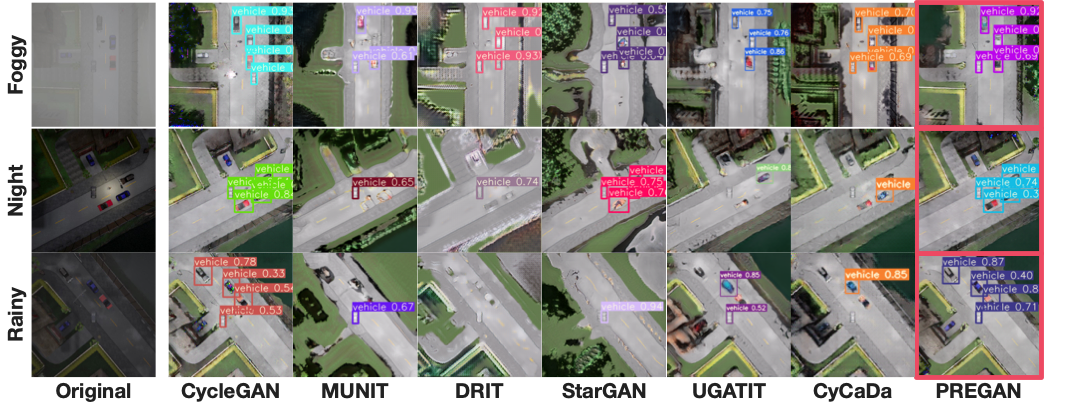}
    \vspace{-5pt}
    \caption{\textcolor{black}{Qualitative comparison of object detection boosting with competing methods in \textit{WeakPair(S)} dataset.}}
    \label{fig:Detection}
    \vspace{-5pt}
    \end{figure*}
    
    \begin{figure*}[t]
        \centering
            \includegraphics[width=0.7\textwidth]{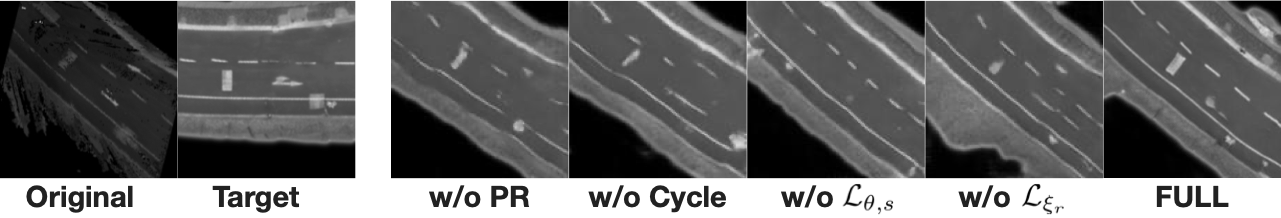}
            \vspace{-3pt}
       \caption{\textcolor{black}{Visual comparison for ablation studies on Stereo $\rightarrow$ Aerial scene in \textit{WeakPair(R)} dataset.}}\label{fig:Ablation}
    \vspace{-10pt}
    \end{figure*}
As shown in TABLE \ref{table: WP Seg}, our method outperforms all the baselines in terms of boosting the road segmentation in the \textit{WeakPair(S)} dataset. While our method is on par with the ground truth(sunny weather) in terms of performance, the \textsc{CycleGAN}, \textsc{StarGAN} and \textsc{MUNIT} is still close behind. This is due to that \textit{WeakPair(S)} dataset is shot from the same kind of sensors and requires less style changes other than simple changes in color and thus requires less disentanglement. 

\noindent \textbf{\textit{WeakPair(R)} \textsc{Dataset}:} With the multi-agent collaborative exploration tasks involved, real-world applications are essential. Fig. \ref{fig:visualComparison} shows the qualitative results of each competing methods and it indicates that the competing baselines that show good performance with the simulated \textit{WeakPair(S)} dataset degenerate seriously in the real-world dataset where style and geometrical transformation is entangled. TABLE \ref{table: AG Seg} provides the corresponding quantitative comparison. The result denotes that when the style changes becomes drastic or across-sensors(e.g. from LiDAR to Aerial camera) with geometrical transform, PREGAN outperforms all these baselines by a large margin.

    \begin{table}[t]
    \textcolor{black}{\caption{Evaluation results of vehicle detetion on orthomosaic perspective in \textit{WeakPair(S)} dataset. The experiments are conducted on ``Foggy" $\rightarrow$ ``Sunny", ``Rainy" $\rightarrow$ ``Sunny", and ``Night" $\rightarrow$ ``Sunny". ``mAP(0.5)" reports the mean average precision with IOU greater than 0.5, and ``F1" reports the harmonic mean of mAP(0.5) and recall.}
    \label{table: WP Detect}
    \vspace{-1pt}
    \resizebox{\linewidth}{!}{%
    \renewcommand{\arraystretch}{1.1}
    \begin{tabular}{@{}lllllll@{}}
    \toprule
    \multicolumn{1}{c}{\multirow{2}{*}{\textbf{Baselines}}} &
      \multicolumn{2}{c}{Foggy} &
      \multicolumn{2}{c}{Rainy} &
      \multicolumn{2}{c}{Night} \\ \cmidrule(l){2-7}
    \multicolumn{1}{c}{} &
      \multicolumn{1}{c}{mAP} &
      F1 &
      \multicolumn{1}{c}{mAP} &
      F1 &
      \multicolumn{1}{c}{mAP} &
      F1 \\ \cmidrule(r){1-1}
             Sunny(GT)      &   92.90  & 92.94 &  92.90  & 92.94 &  92.90  & 92.94 \\
             Original       &   10.26  & 13.54 &  11.74  & 14.21 &  13.55  & 14.79 \\
    \textsc{CycleGAN}       &   52.14  & 55.67 &  51.96  & \textcolor{blue}{57.28} &  53.71  & 57.92 \\
    \textsc{UGATIT}         &   12.48  & 13.22 &  29.04  & 38.51 &  \textcolor{blue}{69.40}  & \textcolor{blue}{76.87} \\
    \textsc{StarGANv2}      &   50.95  & 52.18 &  37.03  & 42.99 &  46.12  & 42.69 \\
    \textsc{DRIT}           &   39.54  & 39.88 &  15.21  & 13.00 &  49.84  & 49.77 \\
    \textsc{MUNIT}          &   \textcolor{blue}{66.11}  & \textcolor{blue}{69.80} &  42.58  & 57.03 &  55.99  & 59.39 \\
    \textsc{CyCaDa}         &   57.02  & 61.57 &  \textcolor{blue}{52.06}  & 49.25 &  59.33  & 58.80 \\
    \textbf{Ours(PREGAN)}   &   \textcolor{red}{86.60} & \textcolor{red}{87.59} & \textcolor{red}{88.69} & \textcolor{red}{90.39} & \textcolor{red}{88.52} & \textcolor{red}{89.94} \\ \bottomrule
    \end{tabular}%
    }
    \vspace{0pt}}
    \end{table}

\subsection{Object Detection}
\label{subsec:Object Detection}
 Demonstration of the detection result is shown in Fig. \ref{fig:Detection}. It shows that even though \textsc{CycleGAN} is able to provide images of decent quality, it tends to generate vehicles that never exist and also will inevitably generate a random shape in the center of the image. \textsc{MUNIT}, \textsc{DRIT} and \textsc{StarGAN} tend to neglect tiny details which results in the blurring effects in objects and thus failed to outline and paint these vehicles. Quantitative indicators shown in TABLE \ref{table: WP Detect} show these observations above and confirms that in the aspects of retrieving details in weakly-paired and boosting object detection, PREGAN leads by a huge margin.

    \begin{table}[t]
    \textcolor{black}{\caption{Quantitative comparison for ablation studies on Stereo $\rightarrow$ Aerial scene in \textit{WeakPair(R)} dataset. It indicates that the main contributing function is the ``PR" without whom the performance drops immediately. More results are in the Appendix.}
    \label{table:Ablation}
    \resizebox{\linewidth}{!}{%
    \renewcommand{\arraystretch}{1}
    \begin{tabular}{@{}lllllll@{}}
    \toprule
      \multicolumn{1}{c}{\textbf{Ablations}} &
      \multicolumn{1}{c}{Original} &
      \multicolumn{1}{c}{w/o PR} &
      \multicolumn{1}{c}{w/o Cycle} &
      \multicolumn{1}{c}{w/o $\mathcal{L}_{\xi_r}$} &
      \multicolumn{1}{c}{w/0 $\mathcal{L}_{\theta,s}$} &
      \multicolumn{1}{c}{Full PREGAN}\\ \cmidrule(l){1-7}
    \textsc{AP}    & 29.71 &  77.83  & 82.91 &  91.87  & 90.97 & \textcolor{red}{92.06}\\
    \textsc{MaxF1} & 43.04 &  74.82  & 81.66 &  93.67  & 92.35 & \textcolor{red}{94.09}\\ \bottomrule
    \end{tabular}%
    }
    \vspace{-10pt}}
    \end{table}

\subsection{Ablation Study}
\label{subsec:Ablation}

\textcolor{black}{Several experiments are designed for ablation studies and all of which are conducted on the \textit{WeakPair} dataset and evaluated with the performance of road segmentation boosting. Firstly, we validate the roles of two self-supervision $\mathcal{L}_{\theta,s}$ and $\mathcal{L}_{\xi_r}$ in the \textbf{FULL}(denotes as ``w/o $\mathcal{L}_{\theta,s}$" and ``w/o $\mathcal{L}_{\xi_r}$"). The relevant result shown in Fig. \ref{fig:Ablation} indicates that the image generated without either one of the self-supervision leads to ambiguity in detail. Then we further deprecate the cycle-consistency by eliminating cycle loss $\mathcal{L}_{cycle}$ (denotes as ``w/o Cycle''). The visual result indicates that without a cycle loss, the style transferring process can be excessive or insufficient. Finally, we deprecate the Pose Randomization architecture (denotes as ``w/o PR"). The result generated by a ``w/o PR" is totally distorted in road's orientation and detail rehabilitation. The ``w/o PR" is tricked and misled by the highly entangled geometric transformation and fails to concentrate on style learning. The road segmentation performs weakly in all scenarios above with unsatisfying results and the quantitative comparison is shown in TABLE \ref{table:Ablation}. We now reassure that the Pose Randomization plays the most important roll in PREGAN while SS keeps details alive. More quantitative results can be found in the Appendix.}

\section{Conclusion}
\label{sec:Conclusion}
We present an approach for style translation between weakly-paired images who are entangled in style and pose transformations. To achieve this, we disentangle the style and pose with pose randomization and train a GAN network with self-supervision by differentiable pose estimator, namely PREGAN. In various experiments, the proposed PREGAN presents the better performances than other comparative methods, showing its potential for real application.

\printbibliography

\clearpage
\appendix

\subsection{Explanation on Self-supervision}
We researched through successful GAN with self-supervision\cite{gidaris2018unsupervised,chen2019self,tran2019improved}, and their methods are more or less the same: rotating or transforming the images with a limited choice of fixed transformations, and introduces classifier to supervise the transformation. They are all translating images pairs that are aligned in the gravity direction and that the initial poses of the images can be considered as $0^\circ$. When predicting the poses in the self-supervision, they assume that the transformed images does not exist in the dataset. For example, if there exists an image ``B'' in the dataset which happens to the same with an $90^\circ$ rotated ``A'', it is pathological to train a classifier that classifies ``B'' as $0^\circ$ and the rotated ``A'' as $90^\circ$. Weakly-paired data is just the unfortunate case. The disadvantages of these methods are obvious when encountered the weakly-paired images: the fixed form of transformation and classification is deficient in treating the chaotic pose in the weakly-paired images, and they do not make use of the assumption that poses between weakly-paired images is estimable. This leads to the novel self-supervision that we proposed, which both make use of the assumption above and is capable of estimating sufficient randomized pose injection.

The rotation output in DPC estimation, due to its nature in the frequency domain, is capable of indicating whether or not the two image is aligned in style: if the two images are style-aligned and share a valuable overlapped region, the estimated rotation between the two images will be reasonable; vice versa, if they are not similar enough, the phase correlation will not recognize any shared content even if they do have overlapped region, and will give an output randomly picked from $0^\circ$ or $90^\circ$. 

Therefore, the DPC in this paper is a $\mathbb{SIM}(2)$ solver with the capability of back-propagation, and the input style should be the same if we expect a correct output from it. This leads to the usage of DPC based self-supervision in this paper.

Please have a look at the Fig.\ref{fig:netstructure} upon which the following description of self-supervision is elaborated.

The same $Original$ image and $Target$ image are pose injected with the randomized pose
$\bigl(\begin{smallmatrix}\hat{s}_{1}\boldsymbol{R}_{\hat{\theta}_{1}} & \boldsymbol{\hat{t}}_{1}, \\ 0 & 1\end{smallmatrix}\bigr)$ 
with the outcomes of $Original Rand$ and $Target Rand$ respectively. The $Original$ and $Original Rand$ are passed to the generator with the result of $Fake Target$ and $Fake Target Rand$. By now, the prerequisite of the self-supervision is acquired. 

 There are two stages of self-supervised loss in the pipeline:
\begin{itemize}
    \item \textbf{Supervision with KLDivergence loss between $Map_{1}$ and $Map_{2}$.} Heatmap surrounded by purple and green is the phase correlation between $Fake Target$ and $Target$ calculated by DPC, and is denoted as $Map_{1}$. By analogy, Heatmap $Map_{2}$ surrounded by orange and red is from the calculation of $Fake Target Rand$ and $Target Rand$. The stronger the generator is, the more style-aligned  $Fake Target$ and $Fake Target Rand$ are with $Target$ and $Target Rand$, resulting in the consistency of the phase correlation output($Map_{1}$ and $Map_{2}$). If the generator is weak and fails to generate satisfying $Fake Target$ and $Fake Target Rand$, the peak of $Map_{1}$ and $Map_{2}$ will be jumping between 0 degree and 90 degree, which is unstable and there will be a big KLD loss.
    \item \textbf{Supervision with KLDivergence loss between $Map_{3}$ and $Map_{4}$}. Heatmap surrounded by purple and orange is the phase correlation between $Fake Target$ and $Fake Target Rand$ calculated by DPC, and is denoted as $Map_{3}$. $Map_{4}$ is a one-peak tensor with respect to 
    $\bigl(\begin{smallmatrix}\hat{s}_{1}\boldsymbol{R}_{\hat{\theta}_{1}} & \boldsymbol{\hat{t}}_{1}, \\ 0 & 1\end{smallmatrix}\bigr)$.
    This self-supervised loss is designed to help GAN to disentangle style and pose better. If the GAN is weak and is confused by the pose between the two inputs while trying to focus on the style translation, $Fake Target$ and $Fake Target Rand$ will be highly chaotic, and are not simply two translated images with a 
    $\bigl(\begin{smallmatrix}\hat{s}_{1}\boldsymbol{R}_{\hat{\theta}_{1}} & \boldsymbol{\hat{t}}_{1}, \\ 0 & 1\end{smallmatrix}\bigr)$
    pose transformation. Therefore, calculation the loss between $Map_{3}$ and $Map_{4}$ is helpful, and in our case, we calculated the KLDiv loss between $Map_{3}$ and $Map_{4}$.
\end{itemize}

\begin{figure}[t]
        \centering
        \includegraphics[width=\linewidth]{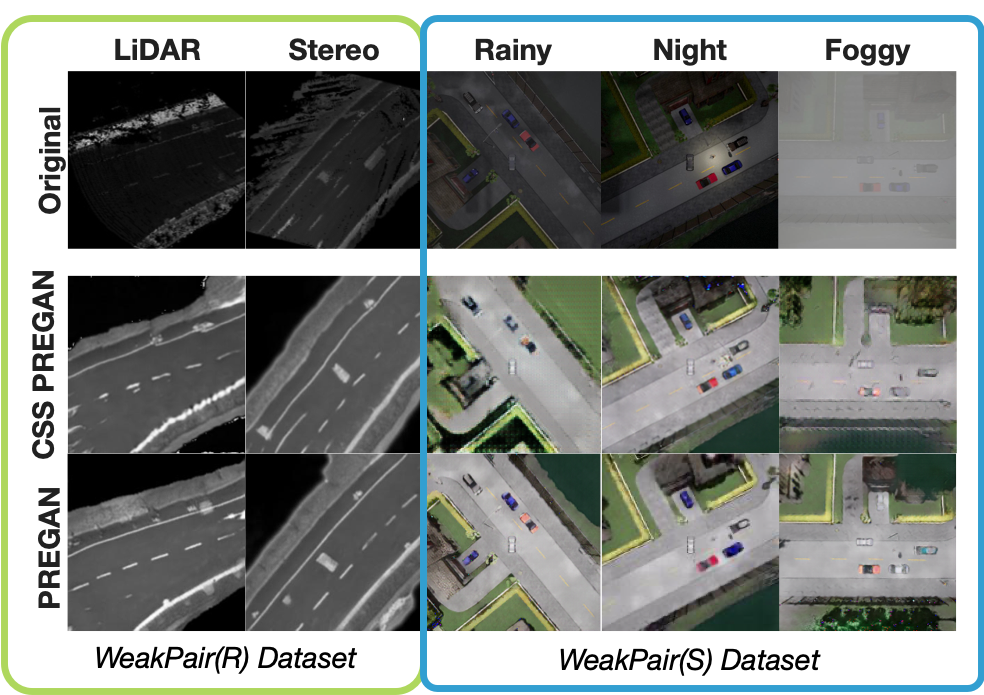}
        \vspace{0pt}
        \caption{Qualitative comparison of different types of self-supervision in \textit{WeakPair(S)} and \textit{WeakPair(R)} dataset. ``CSS PREGAN'' is the FULL version of PREGAN but with the novel self-supervision and pose injection altered by classifier in common practice.}
        \label{fig:visualComparison_ss}
        \vspace{0pt}
\end{figure}

To demonstrate that the self-supervision in common practice is not practical in weakly-paired setting, we conduct experiments by altering our unique self-supervision with the classical classifier mentioned above, and also by changing the randomized pose injection to fixed angle rotations. The qualitative result is shown in Fig.\ref{fig:visualComparison_ss} and the quantitative result of road segmentation boosting is shown in Table.\ref{table:Ablation_ss}.

    \begin{table}[t]
    \caption{Quantitative comparison for ablation studies on Stereo $\rightarrow$ Aerial scene in \textit{WeakPair(R)} dataset.}
    \label{table:Ablation_ss}
    \resizebox{\linewidth}{!}{%
    \renewcommand{\arraystretch}{1}
    \begin{tabular}{@{}llll@{}}
    \toprule
      \multicolumn{1}{c}{\textbf{Ablations}} &
      \multicolumn{1}{c}{Original} &
      \multicolumn{1}{c}{CSS PREGAN} &
      \multicolumn{1}{c}{Full PREGAN}\\ \cmidrule(l){1-4}
    \textsc{AP}    & 29.71 & 80.47 & \textcolor{red}{91.92}\\
    \textsc{MaxF1} & 43.04 & 76.91 & \textcolor{red}{94.49}\\ \bottomrule
    \end{tabular}%
    }
    \vspace{-10pt}
    \end{table}

\subsection{Difference between Pose Injection and Data Augmentation}

For data augmentation in any common practice, the transformation in poses (rotating, cropping, splitting) happens in and only in the data loading process and that the methods can be trained with or without such data augmentation. Another aspect of the data augmentation in style transferring networks is that both the target and the original images are transformed with the same parameter. 

However, inspired by domain randomization which randomizes one domain to make it insensitive, we modeled the weakly-paired problem as the disentanglement of poses and style. Since the purpose is to translate the style regardless of the pose, we randomized the pose with different pose injections in different phases. To make it clear, there are two different types of pose injection in our method: (i)the pose injection is applied to the the original image alone (leaving the target image unchanged) in the ``pose randomization'' process for disentanglement, and (ii) it is applied to both images in the ``self-supervision'' process for generative consistency. This is different from the way in which data augmentation been applied in any common practice. It is also worth mentioning that in the ``BASIC'' version, the pose randomization is not simply a random pose injection, but also requires a differentiable pose estimator for the pose recovery, which separate our method with others. For a better performance and generative consistency, we further developed the ``FULL'' by introducing self-supervision between the images before and after the second pose injection. Operations on such images also separate the method from combinations of data augmentation and self-supervision.

\subsection{Classification on weakly-paired MNIST}

    \begin{figure}[h]
    \centering
    \includegraphics[width=\linewidth]{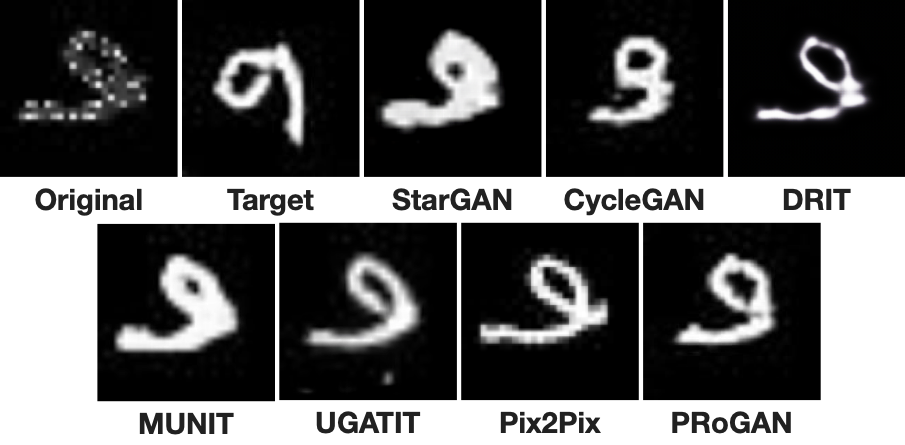}
    \caption{Visual comparison of input image, target image and generated images by each method on \textit{MNIST} dataset. Note: 1) ``Target" is the raw \textit{MNIST} data, 2) the demonstration of \textsc{Pix2Pix} in the figure is the one trained on paired data.}\label{fig:classification}
    \end{figure}

\subsection{Results on Baselines with Data Augmentation}

\begin{figure}[t]
        \centering
        \includegraphics[width=\linewidth]{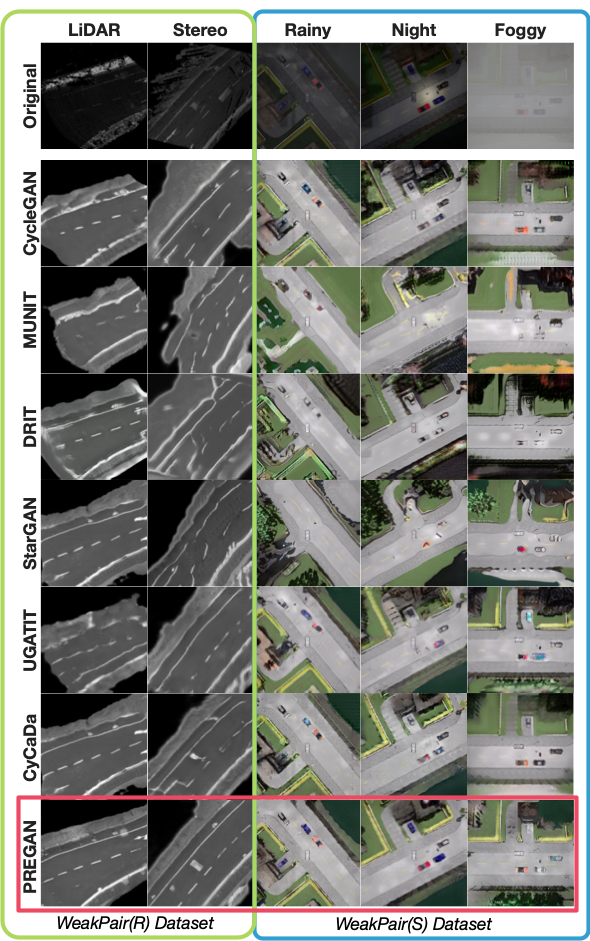}
        \vspace{0pt}
        \caption{Qualitative comparison of competing methods in \textit{WeakPair(S)} and \textit{WeakPair(R)} dataset. All baselines are data augmented by injecting random poses to both of the input image.}
        \label{fig:visualComparison_appendix}
        \vspace{0pt}
\end{figure}

There are two different ways of pose injection in two phases of the PREGAN: (i)the pose injection is applied to the the original image alone (leaving the target image unchanged) in the ``pose randomization'' process for disentanglement, and (ii) it is applied to both images in the ``self-supervision'' process for generative consistency. Even though they are not treated as any ways of data augmentation in the method, we should still apply them to the baselines so that they are in some way given the same data. In the main paper, injecting random poses to the original alone is applied to baselines methods as a way of data augmentation, whereas in the Appendix here, we injecting random poses to both of the inputs to simulate the pose injection in (ii). The results are shown below:

\noindent\textbf{Visual Results:} is shown in Fig.\ref{fig:visualComparison_appendix}

\vfill\break
\vspace{500pt}

\noindent\textbf{Quantitative Results on Road Segmentation Boosting in \textit{WeakPair(S)}:} is shown in Table.\ref{table: WP Seg appendix}

    \begin{table}[h]
    \caption{Evaluation results of road segmentation on orthomosaic perspective in \textit{WeakPair(S)} dataset. The experiments is conducted on ``Foggy" $\rightarrow$ ``Sunny", ``Rainy" $\rightarrow$ ``Sunny", and ``Night" $\rightarrow$ ``Sunny". IoU(percentage) reports the segmentation performance of roads and the higher is better.}
    \label{table: WP Seg appendix}
    \vspace{0pt}
    \resizebox{\linewidth}{!}{%
    \renewcommand{\arraystretch}{1.1}
    \begin{tabular}{@{}lllllll@{}}
    \toprule
    \multicolumn{1}{c}{\multirow{2}{*}{\textbf{Baselines}}} &
      \multicolumn{2}{c}{Foggy(IoU)} &
      \multicolumn{2}{c}{Rainy(IoU)} &
      \multicolumn{2}{c}{Night(IoU)} \\ \cmidrule(l){2-7}
    \multicolumn{1}{c}{} &
      \multicolumn{1}{c}{AP} &
      MaxF1 &
      \multicolumn{1}{c}{AP} &
      MaxF1 &
      \multicolumn{1}{c}{AP} &
      MaxF1 \\ \cmidrule(r){1-1}
             Sunny(GT)  &   92.91 & 97.40 &  92.91 & 97.40 &  92.91  & 97.40 \\
             Original   &   65.73 & 77.54 &  23.26 & 37.74 &  30.17  & 37.65 \\
    \textsc{CycleGAN}   &   89.13 & 88.22 &  89.17 & 89.05 &  87.46  & 91.96\\
    \textsc{UGATIT}     &   87.12 & 85.45 &  82.38 & 82.51 &  88.42  & 91.19 \\
    \textsc{StarGANv2}  &   86.52 & \textcolor{blue}{89.68} &  88.28 & 89.77 &  \textcolor{blue}{88.83}  & \textcolor{blue}{92.79} \\
    \textsc{DRIT}       &   74.66 & 78.80 &  83.74 & 86.53 &  83.07  & 84.21 \\
    \textsc{MUNIT}      &   81.85 & 82.60 &  89.49 & 90.37 &  85.11  & 82.69 \\
    \textsc{CyCaDa}     &   \textcolor{blue}{90.88} & 88.41 &  \textcolor{blue}{90.94} & \textcolor{blue}{90.72} &  86.71  & 88.42 \\
    \textbf{Ours(PREGAN)} & \textcolor{red}{92.15} & \textcolor{red}{94.22} & \textcolor{red}{93.57} & \textcolor{red}{96.59} & \textcolor{red}{92.50} & \textcolor{red}{96.86} \\ \bottomrule
    \end{tabular}%
    }
    \vspace{0pt}
    \end{table}

\noindent\textbf{Quantitative Results on Road Segmentation Boosting in \textit{WeakPair(R)}:} is shown in Table.\ref{table: AG Seg appendix}

    \begin{table}[h]
    \caption{Evaluation results on \textit{WeakPair(R)} Dataset of road segmentation. The experiments is conducted both on LiDAR map $\rightarrow$ Drones's Map and on stereo map $\rightarrow$ Drones's Map. IoU(percentage) reports the segmentation performance of roads and the higher is better.}
    \label{table: AG Seg appendix}
    \centering
    \vspace{0pt}
    \resizebox{0.7\linewidth}{!}{%
    \renewcommand{\arraystretch}{1.1}
    \begin{tabular}{@{}lllll@{}}
    \toprule
    \multicolumn{1}{c}{\multirow{2}{*}{\textbf{Baselines}}} &
      \multicolumn{2}{c}{S2A(IoU)} &
      \multicolumn{2}{c}{L2A(IoU)} \\ \cmidrule(l){2-5}
    \multicolumn{1}{c}{} &
      \multicolumn{1}{c}{AP} &
      MaxF1 &
      \multicolumn{1}{c}{AP} &
      MaxF1 \\ \cmidrule(r){1-1}
             Aerial(GT)            & 92.46  & 97.71 & 92.46 & 97.71 \\
             Original              & 54.67  & 53.59 & 16.12 & 28.26 \\
    \textsc{CycleGAN}              & 25.72  & 26.88 & 29.35 & 42.89 \\
    \textsc{UGATIT}                & 70.69  & 75.37 & 22.83 & 31.55 \\
    \textsc{StarGANv2}             & \textcolor{blue}{79.28}  & \textcolor{blue}{77.21} & 32.06 & 38.52 \\
    \textsc{DRIT}                  & 51.67  & 52.89 & 20.65 & 32.18 \\
    \textsc{MUNIT}                 & 39.11  & 43.44 & 25.19 & 40.05 \\
    \textsc{CyCaDa}                & 63.14  & 64.29 & \textcolor{blue}{40.77} & \textcolor{blue}{50.63} \\
    \textbf{Ours(PREGAN)} & \textcolor{red}{92.06} & \textcolor{red}{94.09} & \textcolor{red}{92.10} & \textcolor{red}{95.21} \\ \bottomrule
    \end{tabular}%
    }
    \vspace{0pt}
    \end{table}

\noindent\textbf{Quantitative Results on Vehicle Detection Boosting in \textit{WeakPair(S)}:} is shown in Table.\ref{table: WP Detect appendix}

    \begin{table}[h]
    \caption{Evaluation results of vehicle detetion on orthomosaic perspective in \textit{WeakPair(S)} dataset. The experiments are conducted on ``Foggy" $\rightarrow$ ``Sunny", ``Rainy" $\rightarrow$ ``Sunny", and ``Night" $\rightarrow$ ``Sunny". ``mAP(0.5)" reports the mean average precision with IOU greater than 0.5, and ``F1" reports the harmonic mean of mAP(0.5) and recall.}
    \label{table: WP Detect appendix}
    \vspace{0pt}
    \resizebox{\linewidth}{!}{%
    \renewcommand{\arraystretch}{1.1}
    \begin{tabular}{@{}lllllll@{}}
    \toprule
    \multicolumn{1}{c}{\multirow{2}{*}{\textbf{Baselines}}} &
      \multicolumn{2}{c}{Foggy} &
      \multicolumn{2}{c}{Rainy} &
      \multicolumn{2}{c}{Night} \\ \cmidrule(l){2-7}
    \multicolumn{1}{c}{} &
      \multicolumn{1}{c}{mAP} &
      F1 &
      \multicolumn{1}{c}{mAP} &
      F1 &
      \multicolumn{1}{c}{mAP} &
      F1 \\ \cmidrule(r){1-1}
             Sunny(GT)      &   92.90  & 92.94 &  92.90  & 92.94 &  92.90  & 92.94 \\
             Original       &   10.26  & 13.54 &  11.74  & 14.21 &  13.55  & 14.79 \\
    \textsc{CycleGAN}       &   52.00  & 56.07 &  52.17  & \textcolor{blue}{60.39} &  54.56  & 57.28 \\
    \textsc{UGATIT}         &   15.12  & 18.76 &  29.22  & 38.74 &  \textcolor{blue}{68.43}  & \textcolor{blue}{78.09} \\
    \textsc{StarGANv2}      &   52.57  & 51.95 &  37.48  & 43.11 &  47.00  & 45.82 \\
    \textsc{DRIT}           &   37.44  & 43.14 &  19.66  & 12.95 &  50.49  & 52.16 \\
    \textsc{MUNIT}          &   \textcolor{blue}{65.99}  & 64.33 &  41.02  & 58.55 &  58.19  & 60.36 \\
    \textsc{CyCaDa}         &   57.85  & \textcolor{blue}{66.38} &  \textcolor{blue}{56.37}  & 48.91 &  60.23  & 58.15 \\
    \textbf{Ours(PREGAN)}   &   \textcolor{red}{86.60} & \textcolor{red}{87.59} & \textcolor{red}{88.69} & \textcolor{red}{90.39} & \textcolor{red}{88.52} & \textcolor{red}{89.94} \\ \bottomrule
    \end{tabular}%
    }
    \vspace{0pt}
    \end{table}
    
 \begin{table}[ht]
    \caption{Evaluation results of boosting \textsc{SIFT} matching in \textit{WeakPair} dataset. We adopt $\mathcal{N}_{SIFT}$ as the metric with the 50 as the top confidence.}
    \label{table:SIFT}
    \centering
    \resizebox{0.8\linewidth}{!}{%
    \renewcommand{\arraystretch}{1}
    \vspace{-10pt}
    \begin{tabular}{@{}llll|ll@{}}
    \toprule
      \multicolumn{1}{c}{\textbf{Baselines}} &
      \multicolumn{1}{c}{Foggy} &
      \multicolumn{1}{c}{Rainy} &
      \multicolumn{1}{c}{Night} &
      \multicolumn{1}{c}{S2A} &
      \multicolumn{1}{c}{L2A} \\ \cmidrule(l){1-6}
    \textsc{CycleGAN}     &  \textcolor{blue}{32.5}  & 35.1 &  \textcolor{blue}{33.4}  & \textcolor{blue}{25.7} & \textcolor{blue}{33.1} \\
    \textsc{UGATIT}      &  10.9  & 13.3 &  11.2  & 21.5 & 32.8 \\
    \textsc{StarGANv2}    &  13.5  & 13.0 & 12.5   & 14.2 & 14.0\\
    \textsc{DRIT}         &  11.3  & 10.9 & 7.5    & 6.0  & 21.9\\
    \textsc{MUNIT}        &  29.2  & \textcolor{blue}{39.5} & 11.7   & 23.5 & 28.1  \\
    \textbf{Ours(PREGAN)} &   \textcolor{red}{44.8} & \textcolor{red}{45.3} & \textcolor{red}{45.9} & \textcolor{red}{45.2} & \textcolor{red}{46.8} \\ \bottomrule
    \end{tabular}%
    }
    \vspace{-10pt}
    \end{table}
\subsection{Results on Feature Matching for Localization}

\begin{figure*}[h]
        \centering
            \includegraphics[width=0.7\linewidth]{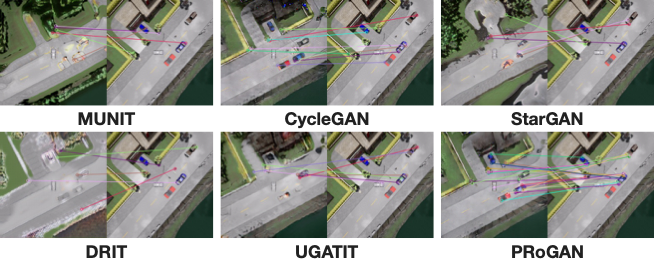}
            \vspace{0pt}
        \caption{Visual comparison of localization with the feature extractor of \textsc{SIFT}.}
        \label{fig:SIFT}
        \vspace{-10pt}
    \end{figure*}

We perform an aerial-ground localization in this experiment by the means of \textsc{SIFT}. The experiments are conducted on the \textit{WeakPair} dataset and therefore the target image is either a ``sunny day shot" or an ``aerial view" respectively. TABLE \ref{table:SIFT} reports the quantitative comparison while Fig. \ref{fig:SIFT} provides a qualitative one. They indicate that our PREGAN obtains the best result on both the adverse weather scenario and the sensor alternation scenario. This experiment assures that our method can preserve more details consistently, thus resulting in more correct feature correspondences.

\subsection{More Quantitative Results on Ablation}

To demonstrate the ablation study better, we have also tested all the ablations with more tasks boosting, e.g. road segmentation and vehicle detection, in different settings. The quantitative results for road segmentation are shown in Table.\ref{table: AblationMoreSeg}, and vehicle detection in Table.\ref{table: AblationMoreDetect}.

    \begin{table}[ht]
    \caption{Quantitative results on the task of road segmentation boosting in different environments for ablation studies.}
    \label{table: AblationMoreSeg}
    \vspace{-1pt}
    \resizebox{\linewidth}{!}{%
    \renewcommand{\arraystretch}{1.1}
    \begin{tabular}{@{}lllllll@{}}
    \toprule
    \multicolumn{1}{c}{\multirow{2}{*}{\textbf{Ablations}}} &
      \multicolumn{2}{c}{Foggy} &
      \multicolumn{2}{c}{Night} &
      \multicolumn{2}{c}{Stereo} \\ \cmidrule(l){2-7}
    \multicolumn{1}{c}{} &
      \multicolumn{1}{c}{mAP} &
      F1 &
      \multicolumn{1}{c}{mAP} &
      F1 &
      \multicolumn{1}{c}{mAP} &
      F1 \\ \cmidrule(r){1-1}
    Original                    & 20.88  & 35.69 & 27.96 & 33.81 &  29.71 & 43.04 \\
    w/o PR                      & 88.72  & 88.47 & 87.95 & 89.39 &  77.83 & 74.82 \\
    w/o Cycle                   & 91.68  & 92.29 & 88.55 & 89.09 &  82.91 & 81.66 \\
    w/o $\mathcal{L}_{\xi_r}$   & 91.87  & 91.55 & 90.40 & 93.06 &  91.87 & 93.67 \\
    w/o $\mathcal{L}_{\theta,s}$& 91.73  & 92.04 & 91.36 & 93.89 &  90.97 & 92.35 \\
    \textbf{Full PREGAN)}       & \textcolor{red}{92.15}  & \textcolor{red}{94.22} & \textcolor{red}{92.50} & \textcolor{red}{96.86} &  \textcolor{red}{92.06} & \textcolor{red}{94.09} \\ \bottomrule
    \end{tabular}%
    }
    \vspace{0pt}
    \end{table}
    
    \begin{table*}[t]
    \caption{Quantitative results on the task of vehicle detection boosting in different environments for ablation studies.}
    \label{table: AblationMoreDetect}
    \vspace{-1pt}
    \centering
    \resizebox{0.5\textwidth}{!}{%
    \renewcommand{\arraystretch}{1.1}
    \begin{tabular}{@{}lllllll@{}}
    \toprule
    \multicolumn{1}{c}{\multirow{2}{*}{\textbf{Ablations}}} &
      \multicolumn{2}{c}{Foggy} &
      \multicolumn{2}{c}{Rainy} &
      \multicolumn{2}{c}{Night} \\ \cmidrule(l){2-7}
    \multicolumn{1}{c}{} &
      \multicolumn{1}{c}{mAP} &
      F1 &
      \multicolumn{1}{c}{mAP} &
      F1 &
      \multicolumn{1}{c}{mAP} &
      F1 \\ \cmidrule(r){1-1}
    Original                    &   10.26  & 13.54 &  11.74  & 14.21 &  13.55  & 14.79 \\
    w/o PR                      & 66.41  & 65.77 & 65.51 & 69.39 &  62.00 & 63.14 \\
    w/o Cycle                   & 70.12  & 71.58 & 70.59 & 73.22 &  69.87 & 73.09 \\
    w/o $\mathcal{L}_{\xi_r}$   & 82.64  & 81.25 & 80.77 & 84.99 &  81.31 & 86.73 \\
    w/o $\mathcal{L}_{\theta,s}$& 83.32  & 80.27 & 82.45 & 82.38 &  83.55 & 83.16 \\
    \textbf{Full PREGAN)}      &\textcolor{red}{86.60} & \textcolor{red}{87.59} & \textcolor{red}{88.69} & \textcolor{red}{90.39} & \textcolor{red}{88.52} & \textcolor{red}{89.94} \\ \bottomrule
    \end{tabular}%
    }
    \vspace{0pt}
    \end{table*}

\end{document}